\documentclass{article}
\pdfoutput=1
\usepackage[utf8]{inputenc}
\usepackage[T1]{fontenc}
\usepackage[english]{babel}
\usepackage{amsmath}
\usepackage{amssymb,amsfonts,textcomp}
\usepackage{array}
\usepackage{hhline}
\usepackage[pdftex]{graphicx}
\begin{document}
\title{Benchmarking of LSTM Networks}
\author{Thomas M. Breuel \\ Google, Inc. \\ {\tt tmb@google.com}}
\date{}
\maketitle

\begin{abstract}
LSTM (Long Short-Term Memory) recurrent neural networks 
have been highly successful in a number
of application areas. This technical report describes the use of the MNIST
and UW3 databases for benchmarking LSTM networks and explores the effect
of different architectural and hyperparameter choices on performance.
Significant findings include: (1) LSTM performance depends smoothly
on learning rates, (2) batching and momentum has no significant
effect on performance, (3) softmax training outperforms least square
training, (4) peephole units are not useful, (5) the standard 
non-linearities (tanh and sigmoid) perform best, (6) bidirectional training
combined with CTC performs better than other methods.
\end{abstract}

\section{Introduction}

LSTM networks \cite{gers2001lstm,gers2003learning,graves2005bidirectional}
have become very popular for many sequence classification
tasks. 
This note presents the results
of large scale benchmarking with a wide range of parameters to
determine the effects of learning rates, batch sizes, momentum,
different non-linearities, and peepholes. The two datasets used for
benchmarking are MNIST and the UW3 text line OCR task.
The questions we are addressing are:
\begin{itemize}
\item Generally, how do LSTMs behave for different hyperparameters?
\item How reproducible are training results based on hyperparameters?
\item What are the effects of batching and momentum on error rates?
\item How do different choices of non-linearities affect performance?
\item Are peepholes useful?
\item What are the effects of bidirectional methods?
\item What are the effects of using CTC?
\end{itemize}

\begin{figure}[tp]
\centerline{ \includegraphics[width=\textwidth,height=2.5in,keepaspectratio]{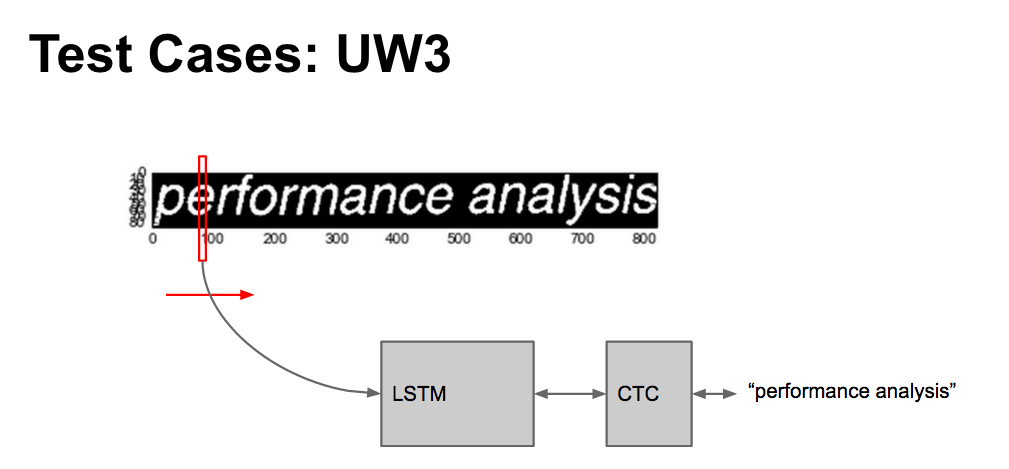} }
\caption{\label{lstm-uw3}
Application of LSTMs to OCR.
}
\end{figure}

\section{Input Data}

We use two kinds of input data in our experiments: MNIST \cite{lecun1998mnist} 
and UW3 \cite{uw3}. 
MNIST is a widely used benchmark on isolated digit handwriting classification.
UW3 is an OCR evaluation database.

We transform both the MNIST and the UW3 inputs into a sequence classification problem
by taking the binary image input and scanning it left to right using vertical slices to the image.
MNIST images are 28x28 pixels large, so this yields a sequence of 28 bit vectors of dimension 28.
UW3 images are variable size, but they are size-normalized and deskewed to a height of 48 pixels;
they still have variable width.

\begin{figure}[tp]
\centerline{ \includegraphics[width=\textwidth,height=2.5in,keepaspectratio]{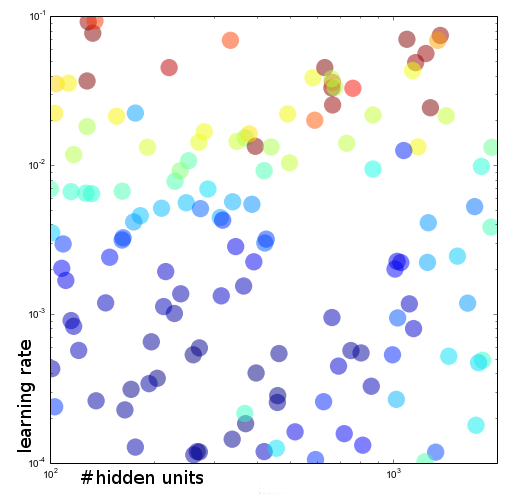} }
\caption{\label{mnist-params}
The parameter space explored during MNIST training of LSTMs. 
Error rates are indicated by color, representing a range of approximately 0.8\% to about 2\% error.
Note that performance
is quite consistent across the parameter space: similar learning rates and numbers of hidden units
give similar error rates. 
}
\end{figure}

\begin{figure}[tp]
\centerline{ \includegraphics[width=\textwidth,height=2.5in,keepaspectratio]{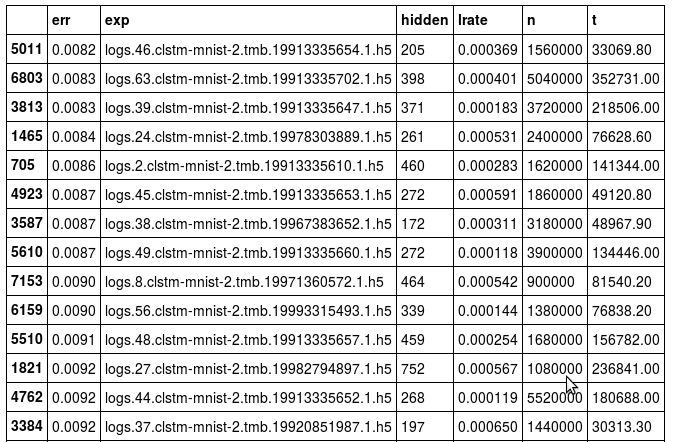} }
\caption{\label{mnist-best-lstm}
These are the best performing networks among the 660 LSTMs trained in
the experiments described in Section~\ref{mnist-by-lr}
}
\end{figure}

\section{MNIST Performance by Learning Rates and Network Size}
\label{mnist-by-lr}

In the initial experiments, LSTM models were trained on MNIST data with between 50 and 500 states and learning rates between $10^{-6}$ and $10^{-1}$.
Figure~\ref{mnist-params} shows the performance generally across different combination of number of states and learning rates.
The figure shows that test set error rate depends quite smoothly on hyperparameters.
As in other neural network learning models, training diverges above some upper limit for the learning rate.
Error rates also increase for large numbers of hidden units and low learning rates.
This is mostly due to learning being very slow, not overtraining.
A look at the top 10 test set error rates (Figure~\ref{mnist-best-lstm}) shows that it is fairly easy to achieve error rates between 0.8\% and 0.9\%.

%
%
%

\begin{figure}[tp]
\hbox{
\centerline{ \includegraphics[width=\textwidth,height=2.5in,keepaspectratio]{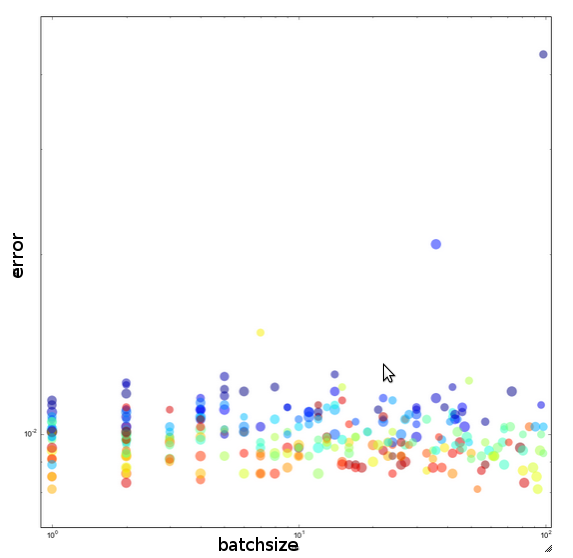}~~~\includegraphics[width=\textwidth,height=2.5in,keepaspectratio]{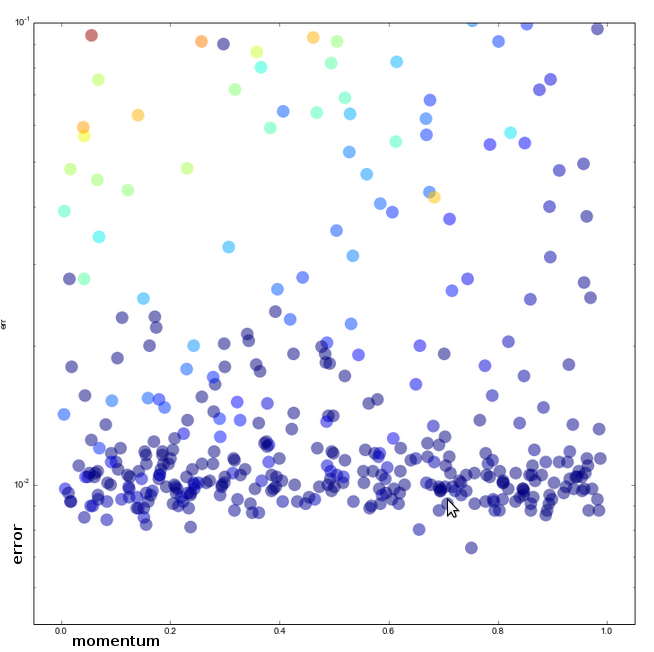} }
}
\caption{\label{mnist-batchsize}
LSTM performance on MNIST is approximately independent of batchsize and
momentum.
}
\end{figure}

\begin{figure}[tp]
\centerline{ \includegraphics[width=\textwidth,height=2.5in,keepaspectratio]{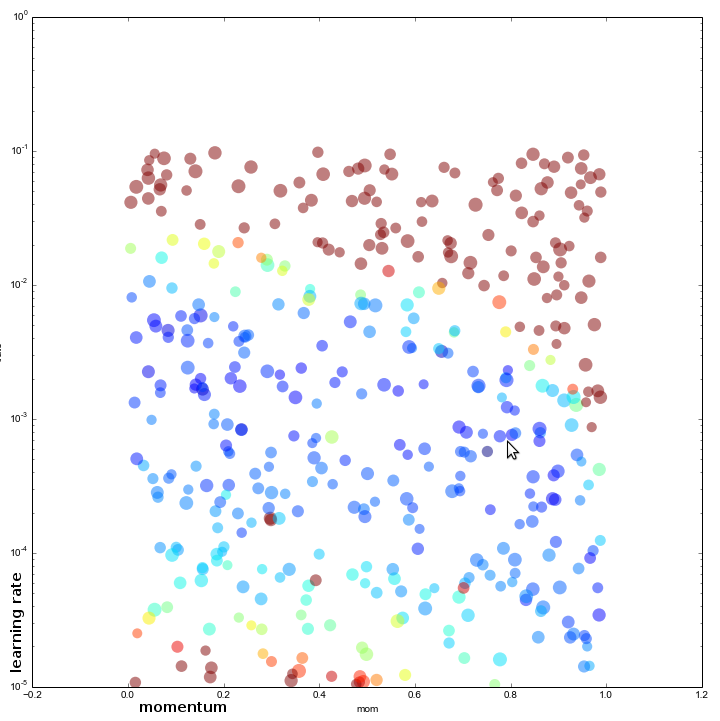} }
\caption{\label{mnist-momentum-params}
The optimal learning rate scales as $\frac{1}{1-\mu}$ witih momentum $\mu$.
}
\end{figure}

\section{Effects of Batchsize and Momentum}

Stochastic gradient descent often is performed with minibatches, computing
gradients from a small set of training samples rather than individual samples.
Such methods are supposed to ``smooth out'' the gradient.
They also allow greater parallelization of the SGD process.
Closely related to minibatches is the use of {\em momentum}, which effectively
also averages gradients over multiple training samples.

For regular neural networks trained with stochastic gradient descent, batch
sizes interact in complex ways with learning rates and nonlinearities.
In particular, for sigmoidal nonlinearities, beyond a certain batch size,
the learning rate needs to be scaled by the inverse of the batch size
in order to avoid divergence;
as a secondary effect, large batch sizes generally fail to achieve the same
minimum error rates that single sample updates achieve.
For ReLU (rectified linear, $f(x) = \max(0, x)$)
nonlinearities, we don't observe the same kind of batchsize dependencies,
however.

To test whether batch size dependencies exist for LSTM networks, 427 networks
were trained with batch sizes ranging from 20 to 2000, momentum parameters between 0 and 0.99, and learning rates between $10^{-5}$ and $10^{-1}$. The results are shown
in Figures~\ref{mnist-batchsize} and~\ref{mnist-momentum-params}. These results show
that there is no significant effect of either batch size or momentum parameter on
error rates.

In addition, for a momentum parameter $\mu$, the optimal learning rate is seen
to scale as $\frac{1}{1-\mu}$; the reason is that with momentum, the same sample
contributes to the update of the gradient effectively that many times.

We note that in these experiments, we obtained the best performance of LSTM
networks on MNIST, with a test set error rate of 0.73\%.


\begin{figure}[tp]
\centerline{ \includegraphics[width=\textwidth,height=2.5in,keepaspectratio]{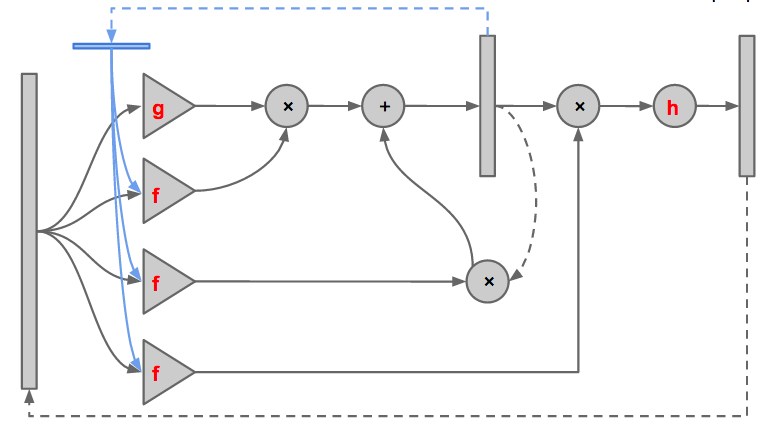}}
\caption{\label{mnist-lstm-structure}
The structure of the LSTM model. The model involves three different
choices of nonlinearities ($f$, $g$, and $h$), plus peephole connections (blue).
Different LSTM structures used in the experiments: 
LINLSTM, LSTM, NPLSTM, RELU2LSTM, RELULSTM, RELUTANHLSTM.
}
\end{figure}

\begin{figure}[tp]
\centerline{\includegraphics[width=\textwidth,height=4in,keepaspectratio]{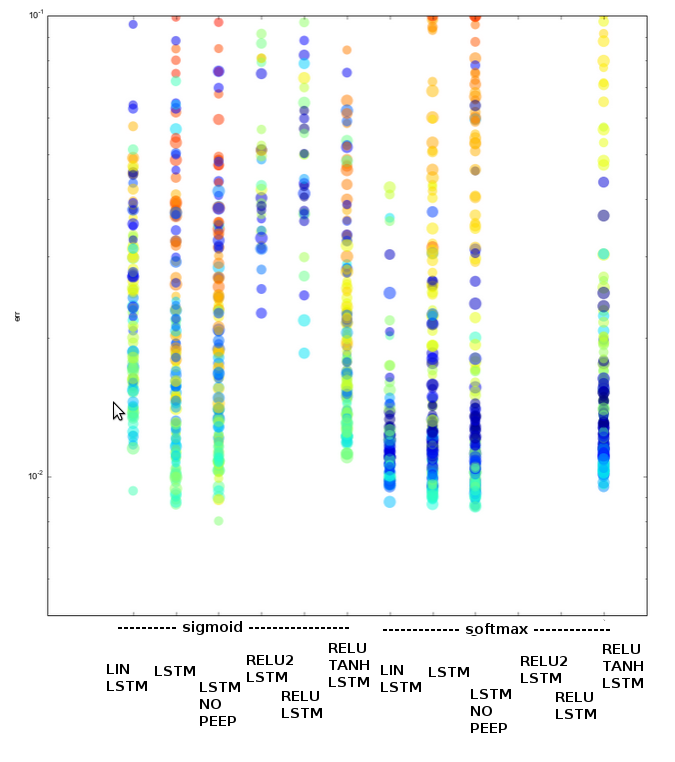}}
\caption{\label{mnist-lstm-types}
The distribution of error rates by LSTM type.
The vertical axis represents error, the horizontal axis LSTM type.
Each dot is one LSTM results, the best test error achieved during a
training run. Dot size indicates number of hidden units and dot color
indicates learning rate.
The left six models use sigmoidal outputs and mean-squared error
at the output, while the right six models use a softmax layer
for output.
Within each group, the results are, from left to right, for
LINLSTM ($h$ = linear), 
LSTM, 
NPLSTM, 
RELU2LSTM ($g$, $h$ = ReLU), 
RELULSTM ($g$ = ReLU, $h$ = linear), 
RELUTANHLSTM ($g$ = ReLU, $h$ = $\tanh$).
The best performing variant is NPLSTM, with either logistic or softmax outputs.
}
\end{figure}

\begin{figure}[tp]
\centerline{\includegraphics[width=\textwidth,height=4in,keepaspectratio]{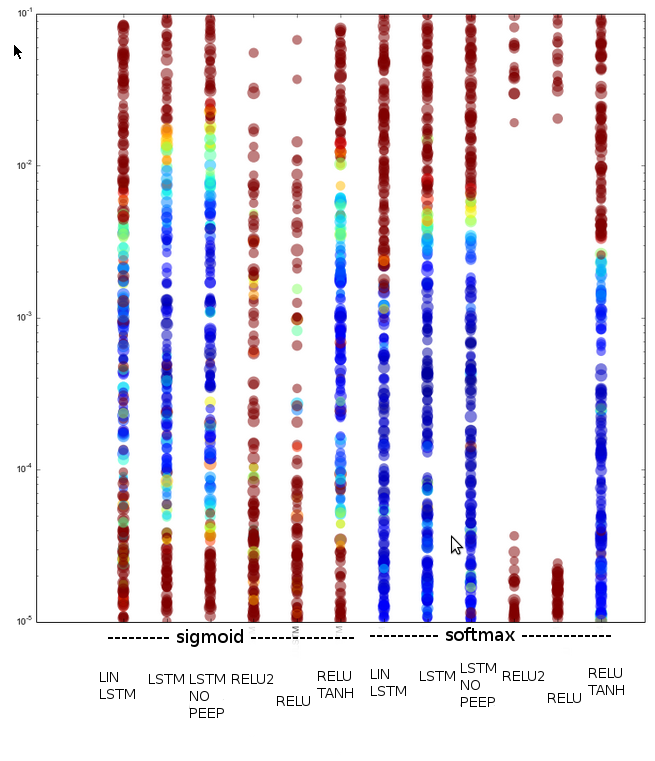}}
\caption{\label{mnist-lstm-types-lr}
Error rates (color) by type and learning rate. The figure verifies that
the range of error rates tried covers the upper range
of convergent learning rates. (RELU2LSTM and RELULSTM both have a gap in
learning rates where the networks all diverged within the first epoch.)
}
\end{figure}

\section{Different LSTM Types applied to MNIST}

LSTM networks involve a number of choices of non-linearities and architecture.
These are illustrated in Figure~\ref{mnist-lstm-structure}.
For regular deep neural networks, we had observed that logistic and softmax output
layers give significantly different results and wanted to see whether
that carries over to LSTM.
We had observed that peephole connections may not help recognition in preliminary
experiments and wanted to verify this.
ReLU nonlinearities appear to give significantly better results on other neural networks, so we investigated whether they also work in the context of LSTMs.

To test these ideas experimentally, 2101 LSTM networks in 12 different configurations
were trained with learning rates between $10^{-6}$ and $10^{-1}$.
The results from these experiments are shown in Figure~\ref{mnist-lstm-types}.
From these results, we can draw the following conclusions:
\begin{itemize}
\item Peepholes do not seem to have a significant effect on error rate.
\item Logistic vs softmax outputs makes no significant difference.
\item Variants with linear outputs or ReLU units perform much worse.
\end{itemize}

Overall, the standard LSTM architecture without peephole connections seems to be
a good choice based on these results.


\clearpage

\begin{figure}[tp]
\centerline{\includegraphics[width=\textwidth,height=4in,keepaspectratio]{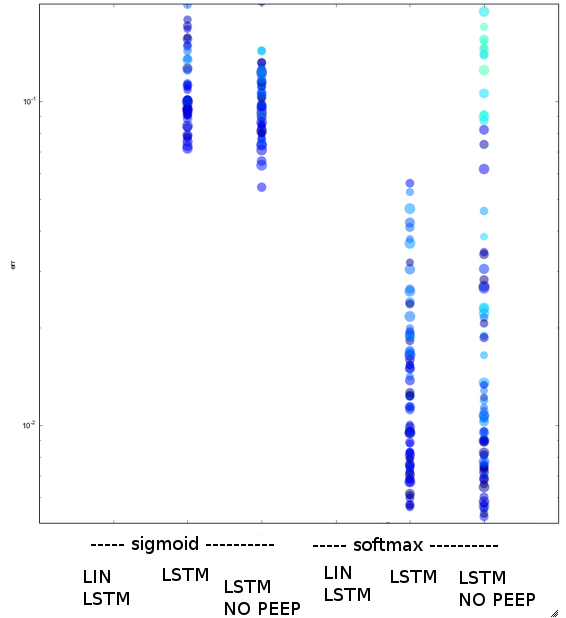} }
\caption{\label{uw3-lstm-types}
Error rates on the UW3 task for different network types.
The left three networks are logistic output variants, the right three are
softmax output variants.
The LSTM types within each group are LINLSTM, LSTM, and NPLSTM.
Unlike MNIST, in these experiments, logistic outputs perform much worse
than softmax outputs.
Networks without peephole connections perform slightly better than networks
containing such connections.
}
\end{figure}

\begin{figure}[tp]
\centerline{\includegraphics[width=\textwidth,height=3in,keepaspectratio]{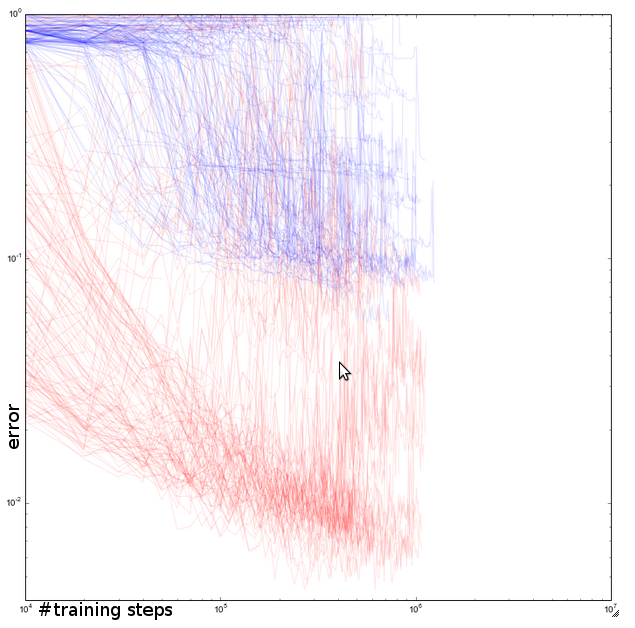}}
\caption{\label{uw3-lstm-types-curves}
Training curves for UW3 LSTM training using logistic output units (blue) and softmax outputs (red).
}
\end{figure}


\section{Different LSTM Types applied to OCR}

The experiments with different LSTM types were also carried out on the UW3 input data
to verify the results on a more complex task.
The biggest difference betweeh MNIST and UW3 is that the input and output sequences
are variable size in UW3, there there are an order of magnitude more class labels, 
and that classes are highly unbalanced. The ReLU variants were not tested
on this dataset.

The most striking difference between UW3 and MNIST results is that on UW3,
logistic outputs perform much worse than softmax outputs.
We verified that in all cases, the range of learning rates covered the
region of convergence.
If we look at the training curves for logistic vs softmax output, we see
that with logistic output units, training starts off slower and ``gets stuck''
on a number of distinct plateaus.
The cause of this still remains to be investigated.

\begin{figure}[tp]
\centerline{ \includegraphics[width=\textwidth,height=2.5in,keepaspectratio]{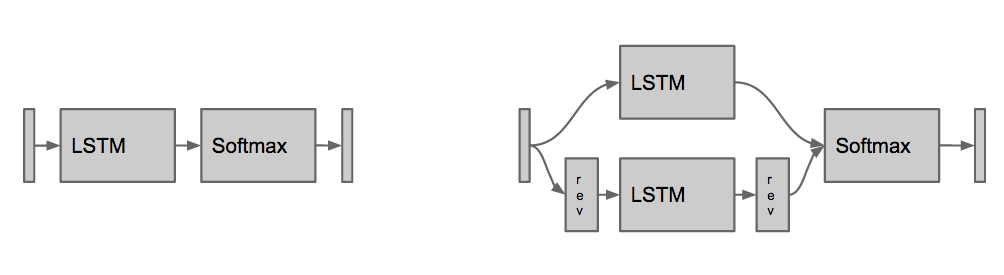} }
\caption{\label{uni-vs-bidi}
LSTM networks can be combined into various more complex network architectures.
The two most common architectures are shown above: unidirectional LSTM (left) 
and bidirectional LSTM (right).
}
\end{figure}

\begin{figure}[tp]
\hbox{
(a)~\includegraphics[width=2in,height=2in,keepaspectratio]{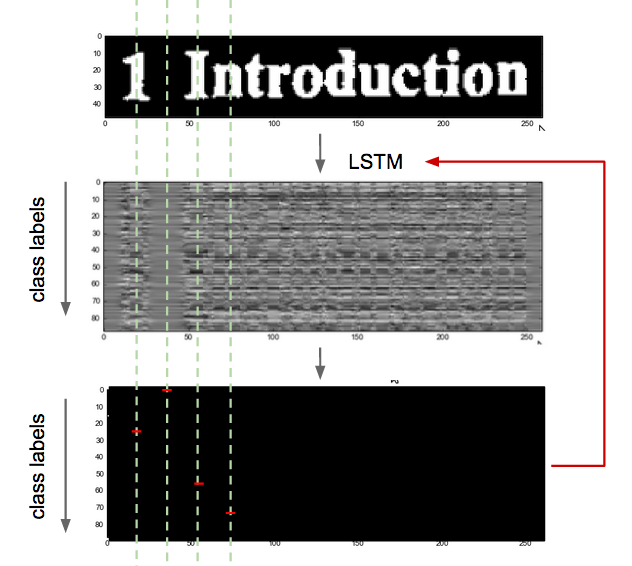} 
(b)~\includegraphics[width=2.5in,height=2in,keepaspectratio]{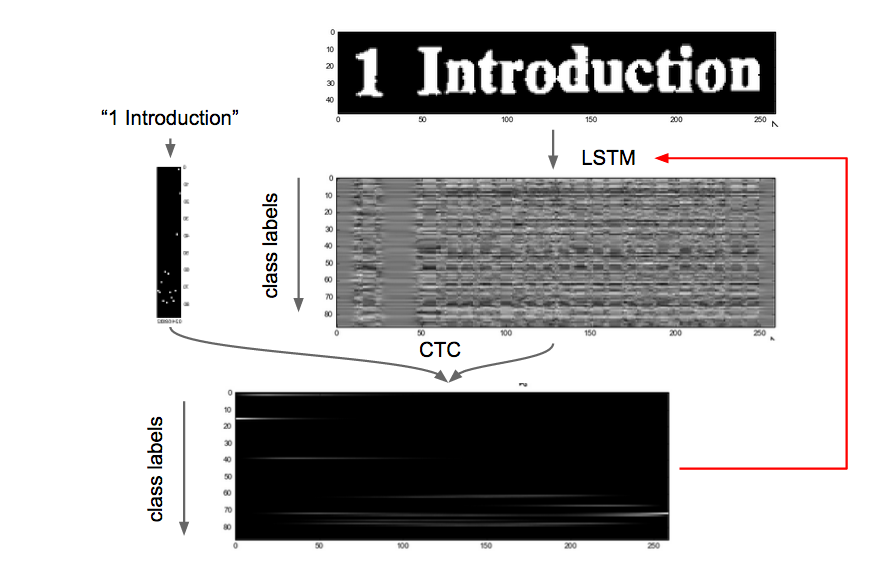} 
}
\caption{\label{ctc-explanation}
In non-CTC training, the target for LSTM training is constructed by assuming a fixed
relationship between the images of input symbols and the apperance of the symbol
in the output sequence.
In CTC training, the target for LSTM training is constructed by aligning (using the
forward-backward algorithm) the target sequence with the actual output of the
LSTM.
}
\end{figure}

\begin{figure}[tp]
\centerline{\includegraphics[width=\textwidth,height=3in,keepaspectratio]{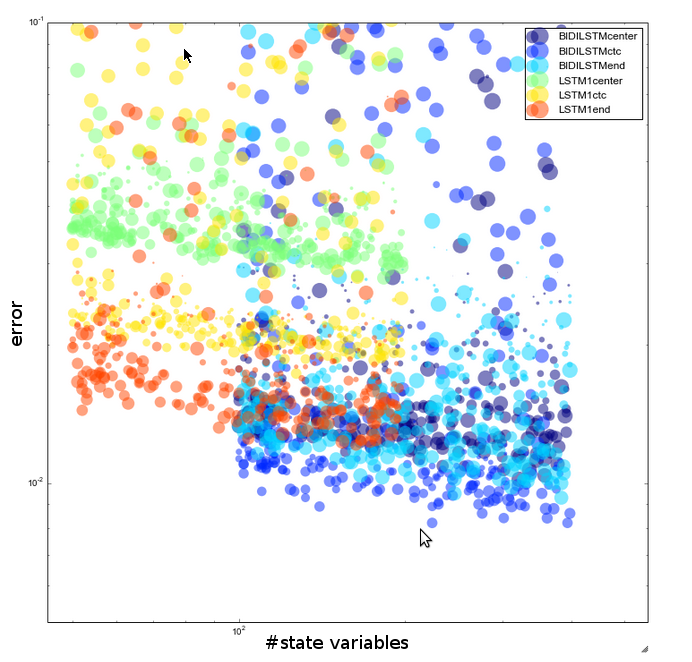}}
\caption{\label{mnist-lstm-unibi-states}
Performance of different network types and training modalities by number of states.
}
\end{figure}

\begin{figure}[tp]
\centerline{\includegraphics[width=\textwidth,height=3in,keepaspectratio]{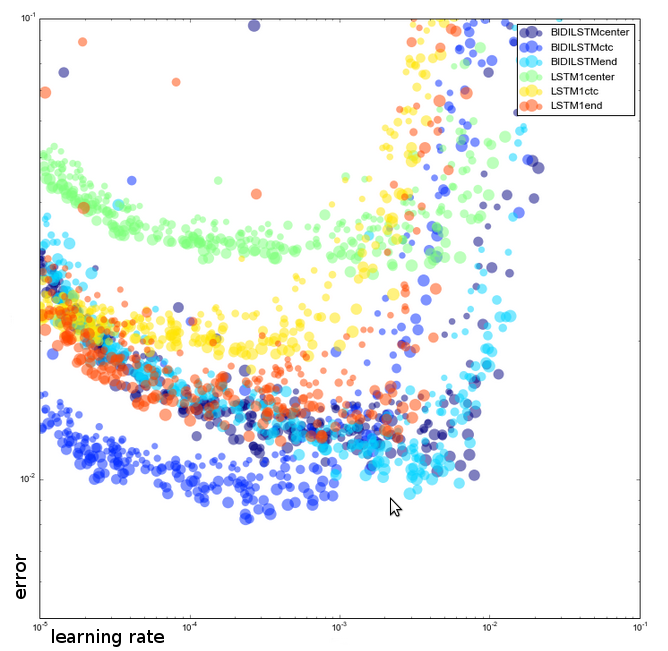}}
\caption{\label{mnist-lstm-unibi}
A plot of error rate by learning rate on MNIST for different network architectures and training modalities.
}
\end{figure}

\begin{figure}[tp]
\centerline{\includegraphics[width=\textwidth,height=3in,keepaspectratio]{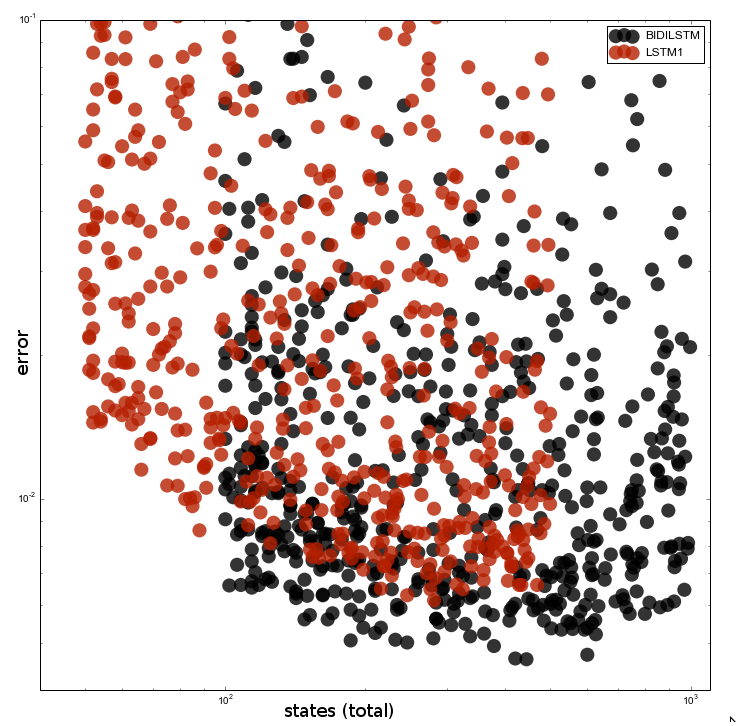}}
\caption{\label{uw3-lstm-unibi-hidden}
Unidirectional vs bidirectional LSTM training on the UW3 database.
}
\end{figure}

\section{Bidirectional vs. Unidirectional, CTC vs. non-CTC}

LSTM networks can be combined into various more complex network architectures.
The two most common architectures are shown above.
The first uses a single LSTM layer followed by a logistic or softmax output layer.
For bidirectional LSTM training, the input sequence is processed both in forward and
reverse, and the combined outputs of the forward and reverse processing at each time
step are then combined into a final output.
Bidirectional LSTM networks can take into account both left and right context in
making decisions at any point in the sequence, but they have the disadvantage that they
are not causal and cannot be applied in real time.
Note that bidirectional networks have twice the number of internal states and slightly
more than twice the number of weights than corresponding unidirectional networks 
with the same number of states.

LSTM networks learn sequence-to-sequence transformations, where input and output
sequences are of equal lengths.
In tasks like OCR, input signals are transformed into shorter sequences of symbols.
Usually, a transcript $ABC$ is represented as the symbolic output 
$\epsilon^+A^+\epsilon^+B^+\epsilon^+C^+\epsilon^+$, augmenting the original set
of classes by an $\epsilon$ symbol.
The LSTM network then predicts a vector of posterior probabilities in this augmented
set of classes.
The non-$\epsilon$ outputs correspond to the ``location'' of the corresponding symbol, 
but there are many possible choices: the location might be consistently at the beginning, center, or end of the symbol, or it might differ from symbol to symbol.
In non-CTC training, the location of the output symbol is fixed based.
In CTC training, the location of the output symbol is determined via the forward
backward algorithm.

In the experiments, we compare unidirectional (``LSTM1'') and bidirectional (``BIDI'') 
networks. Since MNIST contains only a single output symbol per input, for non-CTC
versions, we simply select a constant column (time step) in the output sequence where
the network needs to output the symbol; the two time steps tested are in the middle 
of the output sequence (``center'') and at the end of the output sequence (``end''). 
In addition, CTC was used witih both kinds of networks (``ctc'').

The results of MNIST benchmarks on these six conditions are shown in 
Figure~\ref{mnist-lstm-unibi-states}. Among BIDI networks, CTC training performed
best. Among LSTM1 networks, placing the label at the end performed best.
Not surprisingly, LSTM1center performed worse, because the unidirectional network
could only take into account half the information from the input image before outputting
a label.
Surprisingly, LSTM1ctc performed worse than LSTM1center; that is, CTC training
performed worse than placing the label explicitly at the end of the sequence
for unidirectional training.

The error rate by learning rates and network type is shown in
Figure~\ref{mnist-lstm-unibi}. 
Surprisingly, the different networks achieve their lowest error rates at
learning rates that are different by more than an order of magnitude.
This can be partially explained by the observation that CTC has a different output
class distribution from non-CTC training (in particular, CTC has more non-$\epsilon$
class labels in the output distribution).
But even among the non-CTC training runs, the optimal learning rates differ
depending on network type and location of the target class in time.

Overall, these results confirm the hypothesis that BIDIctc training yields the
best results.
However, the results also caution us against simple benchmarks, since 
learning rates, network structure, and use of CTC interact in complex,
non-monotonic ways.

For UW3, only bidirectional vs. unidirectional training was compared (since the
consistent assignment of a location to characters in a text line is difficult
to achieve).
The results, shown in Figure~\ref{uw3-lstm-unibi-hidden}, are consistent with
the MNIST results: bidirectional LSTM significantly outperforms unidirectional
LSTM networks at all network sizes.
Note that in light of the MNIST result that CTC performs worse than non-CTC
training with unidirectional training, it is possible that unidirectional training
could be improved with a careful manual choice of target locations.




\begin{figure}[tp]
\centerline{\includegraphics[width=\textwidth,height=4in,keepaspectratio]{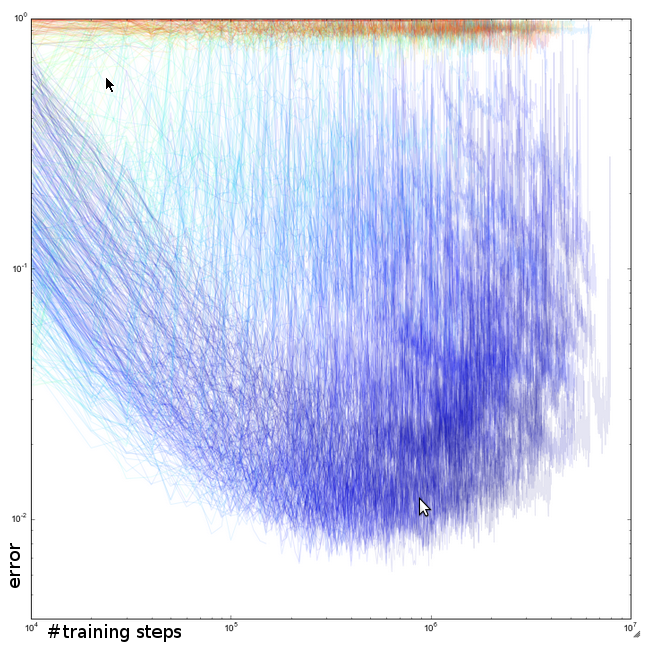}}
\caption{\label{eventual-divergence}
A plot of test-set error vs training steps for unidirectional LSTM training
on the UW3 database. 
Note the logarithmic axes.
Each curve represents one test set error training curve;
the plot represents a total of 942 training runs with a range of learning rates
and number of states.
Color indicates learning rate, from blue (low) to red (high).
}
\end{figure}

\section{Eventual Divergence}

In all experiments where training was continued long enough, we observed eventual
slow divergence of the test set error.
This is shown in Figure~\ref{eventual-divergence}.
Furthermore, the lowest learning rates resulted in the longest time to divergence
This divergence is qualitatively different from the fast divergence we observe
when the learning rates are too high.
Furthermore, it is also represented in the training set error, so it does not
represent overtraining.

Our interpretation of this phenomenon is that LSTM networks internally perform
two separate, competing learning processes.
If we think of an LSTM network as roughly analogous to a Hidden Markov Model (HMM),
these two processes correspond roughly to structural learning and parameter learning.
We postulate that structural learning is a slow process that explores different
structures, with fast parameter learning overlaid on top of this process.
Eventually, (after about one million training steps) in this example, the
network has an optimal structure for the task, and further optimization
results in changes to the structure that are deleterious to overall performance.

\section{Discussion and Conclusions}

Let us summarize the results:

\begin{itemize}
\item LSTM networks give excellent performance on MNIST digit recognition;
    this also represents a simple and useful test case for checking
    whether an LSTM implementation is performing correctly.
\item LSTM network performance is fairly reproducible between training
    runs and test set error has broad flat minima in hyperparameter space
    (i.e., hyperparameter optimization is fairly simple).
\item The best performing networks in all experiments were ``standard''
    LSTM networks with no peephole connections.
\item Peephole connections never resulted in any improved performance.
\item Momentum and batchsize parameters had no observable effect on
    LSTM performance; this means that batching may often be a good
    method for parallelizing LSTM training.
\item LSTM networks failed to converge to low error rate solutions with
    logistic outputs and MSE training for the OCR task; softmax training resulted in the lowest
    error rates overall.
\item CTC and bidirectional networks generally perform better than
    fixed outputs and/or unidirectional networks.
\item LSTM test set error rates seem to invariably diverge eventually.
\end{itemize}

These results agree with other, recently published results on LSTM peformance;
in particular, \cite{greff2015lstm} also found that standard LSTM architectures
with the usual nonlinearities perform best, and that peephole connections do
not improve performance. 
The other results reported above have not been previously obtained.

Numerous other issues remain to be explored experimentally. For example, we do
not know what effect different choices of weight initialization have. Also,
a number of other LSTM-like architectures have been proposed.

We believe that for exploring and benchmarking such architectural variants,
the use of the MNIST and size-normalized UW3 data sets as used in this technical
report form a good basis for comparison, since they are sufficiently difficult
datasets to be interesting, yet still fairly easy to train on.

\section*{Appendix}

The source code used in the experiments is available from {\tt
http://github.com/tmbdev/clstm}.  The MNIST and UW3-derived datasets
are available from {\tt http://tmbdev.net} (in HDF5 format).

\bibliography{lstmbench}
\bibliographystyle{plain}

\end{document}